\documentclass[conference]{IEEEtran}
\IEEEoverridecommandlockouts
\usepackage{cite}
\usepackage{amsmath,amssymb,amsfonts}
\usepackage{algorithmic}
\usepackage{graphicx}
\usepackage{textcomp}
\usepackage{multirow}
\usepackage{xcolor}
\usepackage{hyperref}
\usepackage{soul}

\def\BibTeX{{\rm B\kern-.05em{\sc i\kern-.025em b}\kern-.08em
    T\kern-.1667em\lower.7ex\hbox{E}\kern-.125emX}}
\begin{document}

\newcommand{\change}{changed}
\newcommand{\nochange}{consistent}

\title{Change of Scenery: Unsupervised LiDAR Change Detection for Mobile Robots
\thanks{All authors are with the University of Toronto Robotics Institute.}
}

\author{\IEEEauthorblockN{Alexander D. Krawciw}
\IEEEauthorblockA{\texttt{alec.krawciw@mail.utoronto.ca} \\
}
\and
\IEEEauthorblockN{Jordy Sehn}
\IEEEauthorblockA{\texttt{jordy.sehn@mail.utoronto.ca}}
\and
\IEEEauthorblockN{Timothy D. Barfoot}
\IEEEauthorblockA{\texttt{tim.barfoot@utoronto.ca}}
}

\maketitle

\begin{abstract}
This paper presents a fully unsupervised deep change detection approach for mobile robots with 3D LiDAR.
In unstructured environments, it is infeasible to define a closed set of semantic classes. 
Instead, semantic segmentation is reformulated as binary change detection.
We develop a neural network, RangeNetCD, that uses an existing point-cloud map and a live LiDAR scan to detect scene changes with respect to the map.
Using a novel loss function, existing point-cloud semantic segmentation networks can be trained to perform change detection without any labels or assumptions about local semantics.
The mean intersection over union (mIoU) score is used for quantitative comparison. 
RangeNetCD outperforms the baseline by 3.8\% to 7.7\% depending on the amount of environmental structure.
The neural network operates at 67.1 Hz and is integrated into a robot's autonomy stack to allow safe navigation around obstacles that intersect the planned path. 
In addition, a novel method for the rapid automated acquisition of per-point ground-truth labels is described. Covering changed parts of the scene with retroreflective materials and applying a threshold filter to the intensity channel of the LiDAR allows for quantitative evaluation of the change detector. 
\end{abstract}


\section{Introduction}
While significant progress has been made towards robot navigation in unstructured environments, challenges with robust perception and terrain assessment remain \cite{wijayathunga_challenges_2023}.
These can be exacerbated by a domain gap where methods designed for one location do not transfer reliably to another \cite{kouw_review_2021}.
On-road autonomous driving benefits from clear rules and common types of objects that assist with risk assessment and detection \cite{ma_rethinking_2022}. 
Using extremely large, labelled datasets such as the Waymo Open Dataset \cite{Sun_2020_CVPR} or SemanticKITTI \cite{behley2019iccv} have allowed deep 3D networks to progress rapidly in the semantic and instance segmentation of known classes \cite{yan20222dpass}, \cite{zhou_panoptic-polarnet_2021}. 

\begin{figure}
    \centering
    \includegraphics[width=0.48\textwidth]{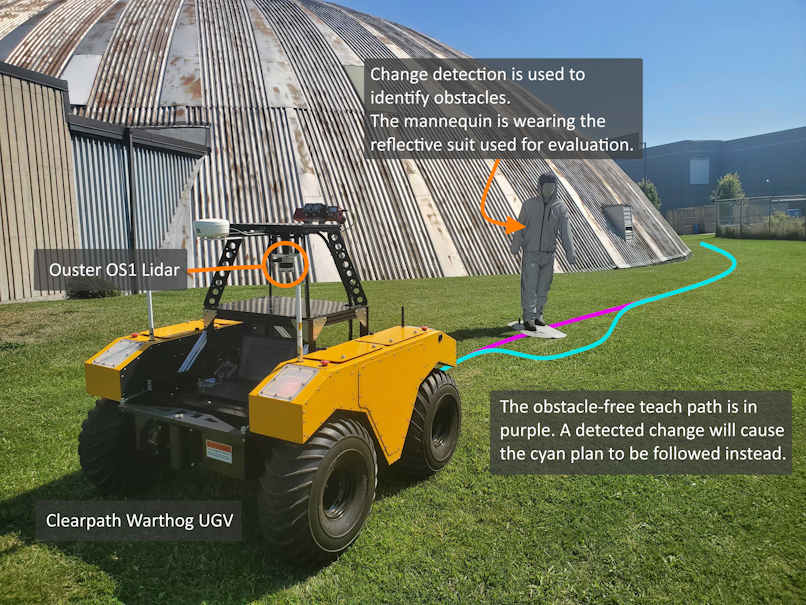}
    \caption{The Clearpath Warthog UGV with the Ouster OS-1 LiDAR, driving a previously taught path. A section of the path is now blocked and the change detection algorithm proposed in this paper will be used to allow the robot to safely navigate around the obstruction. The mannequin is wearing the retroreflective suit that is used for dataset generation and evaluation.}
    \label{fig:coatGrizz}
\end{figure}

This paper aims to solve a broader problem. We envision a situation where a robot can safely navigate autonomously after an initial mapping step. The map could be generated through manual exploration or prior experience from other systems. We argue that a map is a stronger prior for the types of features and obstacles that may be encountered than a predefined set of classes. 
It is desirable to perform per-point segmentation because bounding box regression implicitly imposes class-related sizes. 
Based on these conditions, we formulate the problem as binary change detection; points in the live scan are labelled as either changed or consistent with respect to the map.
A conceptualization of the desired behaviour is provided in \autoref{fig:coatGrizz}.
Classical change-detection methods use nearest-neighbour distances \cite{girardeau2005change}, normal distances \cite{pomerlau_2014}, or ray tracing \cite{underwood_explicit_2013} to classify changes. However, they are most effective indoors and are less accurate in unstructured environments \cite{wu_thesis}. Vegetation is often dynamic on small length scales leading to variable scans with small changes irrelevant to planning. 
We hypothesize that a deep neural network can filter these planning-irrelevant changes to detect higher-quality clusters of changed points. 
Rather than tackle the full scope of terrain traversability assessment \cite{papadakis_terrain_2013}, all changes are treated as a threat and are avoided by the path planner. In the absence of a change, regions where the robot has driven before are considered traversable. 

A novel loss formulation is used to provide an unsupervised training signal. 
This loss leverages inductive biases based on the amount of change, distance of changes from the existing map, and temporal permanence of objects.

These factors can be represented as a standard optimization problem. However, finding the optimal solution is computationally intractable.
Our approach uses a neural network to learn a good heuristic for the problem instead. An unsupervised network training process can be used to learn an initial, general network, which can be fine-tuned to new environments using data collected from the robot itself.

Once the network is trained, it is integrated into an existing autonomy stack to detect obstacles on a Clearpath Warthog unmanned ground vehicle (UGV) \cite{warthog}. The robot can avoid obstacles reliably in real time. 

In summary, we make the following contributions:
\begin{itemize}
    \item an unsupervised deep LiDAR change-detection method for use on mobile robots,
    \item a loss formulation to train neural networks in an unsupervised manner,
    \item a rapid evaluation process for per-point semantic labelling,
    \item a change-detection neural network running in closed-loop on an unmanned ground vehicle. 
\end{itemize}

\begin{figure}[b]
    \centering
    \includegraphics[width=0.46\textwidth]{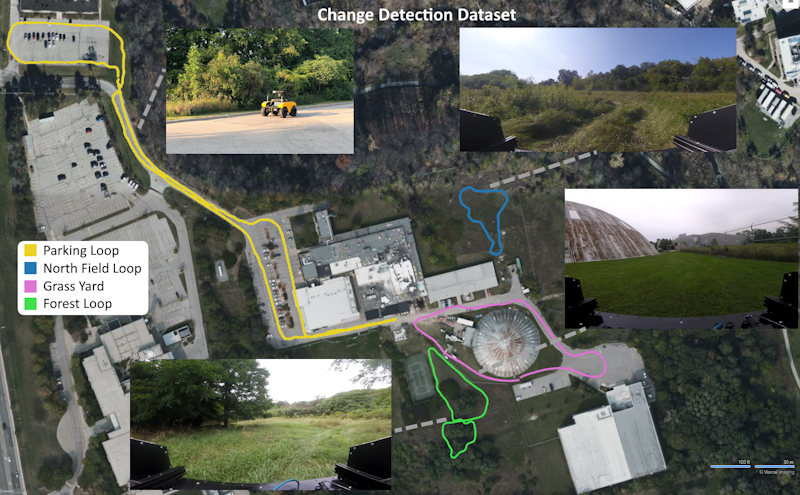}
    \caption{The dataset trajectories and sample views. In total, the dataset is 8.5 km long with 1.75 km of unique paths.}
    \label{fig:dataset}
\end{figure}

The corpus of LiDAR datasets that contain multiple trajectories through the same environment is limited. A custom dataset with a combination of on-road and off-road driving is created and used for evaluation. \autoref{fig:dataset} shows the satellite view of the dataset paths.

\begin{figure*}[tb]
    \centering
    \includegraphics[width=0.85\textwidth]{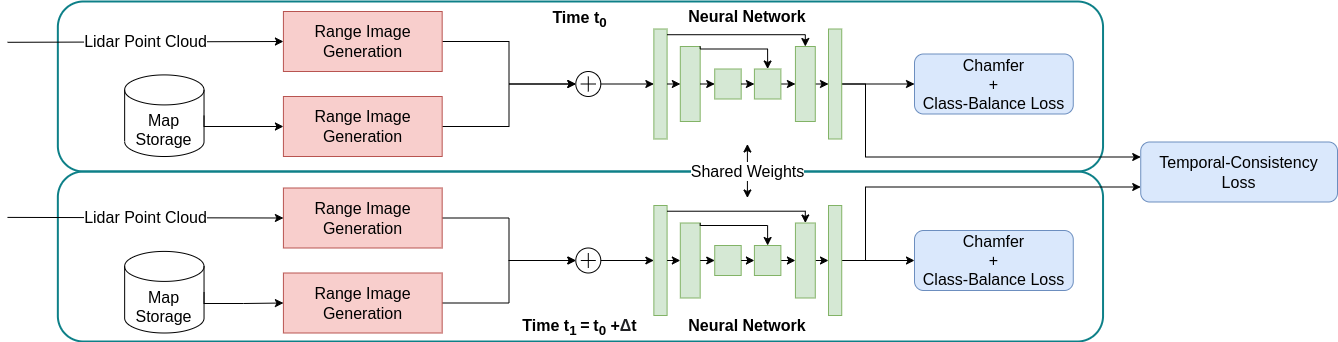}
    \caption{Data flow of the training procedure. Only a single map and live-scan pair are used for inference.}
    \label{fig:blockArch}
\end{figure*}

\section{Related Work}

\subsection{Change Detection}
Change detection in point clouds has been well studied in both mobile robotics \cite{siegwart_cd} and remote sensing \cite{xiao_3d_2023}. 
Classical change detection is primarily based on distances between two scans at different times.
Some works \cite{de_gelis_unsup_2023}, \cite{underwood_explicit_2013} use the labels \textit{Changed} and \textit{Unchanged}, which carry the same meaning as \textit{Changed} and \textit{Consistent} in this work.
Our notation emphasizes that the classifications are not absolute and that points are typically consistent with mapping evidence, not identical.
The simplest method evaluates the distance from each point in the first point cloud to the closest point in the second \cite{girardeau2005change}. 
Distant points are classified as changed if they exceed a threshold value. Wu  \cite{wu_thesis} proposed a Gaussian roughness model of a local neighbourhood to set a dynamic threshold. 
Alternatively, Underwood et al. \cite{underwood_explicit_2013} use ray tracing to determine which parts of the scene have changed. Any points along a ray that are closer than a previous scan are considered changed. A common challenge with these methods is that occluded regions are implicitly assumed to be free space, leading to false positives. 
Voelsen et al. \cite{voelsen_classification_2021} combine these approaches and use region growing to create instance clusters. However, these clusters require a pre-defined set of classes for the changes.

More recently, change detection has been approached using deep learning. There is a focus on airborne LiDARs that detect the long-term change of large geographic areas \cite{OKYAY2019}, \cite{STILLA}. Datasets such as SHREC 2023 \cite{gao_shrec_2023} contain matched pairs of large point clouds for change detection but do not contain trajectory information.
De Gelis et al. \cite{de_gelis_siamese_2023} present SiameseKPConv: a supervised semantic segmentation network that detects six classes of changes on 3D point clouds of cities.
Twin encoders extract features from the map and live-scan point clouds. Once encoded, the feature-space difference of every live point to its nearest map neighbour is used for classification. 
A follow-up work uses the same network architecture but with self-supervised training \cite{de_gelis_unsup_2023}.
They propose three losses: a contrastive loss, a deep clustering loss, and a temporal consistency loss.
The contrastive loss pushes points classified as changed to have different embeddings than those classified as consistent.
By clustering the predictions, some misclassifications are corrected to improve pseudo-labels for the next epoch.
Finally, the temporal consistency loss encourages predictions to be consistent over time, imposing the bias that the number of changed points should be small. This work focuses on large-scale point clouds and does not consider the local trajectory of a mobile system.

Deep learning for change detection on mobile robots is beginning to take shape.
Zhao et al. \cite{zhao_interior_2023} perform supervised change detection on an indoor robot. They use four classes of changes: unchanged, dynamic, structural change, and temporary change.
They show that a range image can be used to make efficient predictions about changes in the environment.
To the best of our knowledge, there are no unsupervised deep LiDAR change detection systems for mobile robots.

\subsection{Map Refinement}
The construction of the point-cloud map impacts change-detection algorithms. Ideally, maps contain only static structures.
Map refinement is the process of updating maps, removing dynamic objects, and rejecting outliers \cite{yguel_update_2008}.
Pioneering works, Erasor \cite{erasor_1} and PeopleRemover \cite{schauer_peopleremoverremoving_2018}, use voxelized pseudo-occupancy-grid maps to estimate whether a region is free at a given time. 
These methods run in batches, allowing information from future scans to improve the detection at every frame. 
Points that exist within voxels that are often empty are removed from the map.
Additionally, Erasor2 \cite{lim_erasor2_2023} uses instance segmentation to reduce the number of misclassifications at contact regions.
Maintenance can be achieved by using multiple passes through the same environment and voting. 

An alternative approach to creating static maps detects dynamic objects while the robot is moving and never adds them to the map.
Pomerleau et al. \cite{pomerlau_2014} use the visibility of the map points and estimate every normal vector to determine whether a point is dynamic or static. 
Static points are added to the SLAM problem and dynamic points are rejected. 
Yoon \cite{yoon_model-free_2019} applies a combination of free-space checks with region growth to detect class-free dynamic points.
Deep learning has also been applied to moving-object segmentation. 
SLIM \cite{andreas_baur_slim_2021} is a self-supervised moving-object segmentation network that operates on point clouds directly. RVMOS \cite{RVMOS} uses sequences of range images and temporal augmentation to train effective supervised classifiers.
Moving-object segmentation alone is insufficient for long-term autonomy because dangerous changes may be static during traversal.

\subsection{Teach and Repeat Framework}
The methods presented in this paper apply to any system that localizes against a point-cloud map. For the purposes of evaluation, we integrate the detector into the Visual Teach and Repeat 3 (VT\&R3) framework\footnote{\href{VT\&R3 Github}{The framework is available at: github.com/utiasASRL/vtr3}}.
VT\&R was designed for stereo cameras and allows robots to precisely localize and repeat a path taught by a human operator \cite{paul2010vtr}.
Wu \cite{wu_thesis} and Sehn \cite{sehn_along_2022} modified the VT\&R3 framework to use LiDAR for odometry and localization. 
During the teaching process, iterative-closest-point (ICP) \cite{burnett_are_2022} odometry determines the motion of the robot. 
These frames are collected into a sequence of topologically connected submaps. A relative transformation between each submap is stored for later localization. 
The repeat phase involves autonomous navigation to any location within the network of paths.
By relaxing the path-following constraint to a corridor \cite{sehn_along_2022}, the robot is given the flexibility to navigate around obstacles on the path. 
A local cost map is required by the existing planner within the corridor around the robot. The proposed method detects changes and treats their locations as unsafe to drive.

\section{Methodology}

\subsection{Problem Definition}
\label{sec:problem_def}
We let the problem of LiDAR change detection be one of binary classification. Given two point clouds in a common reference frame, the map $M$ and current scan $S$, every point $s_i \in S$ is assigned a binary label ($l_i)$ of \textit{Changed} or \textit{Consistent}. Let the number of points in $S$ be $n$. \textit{Consistent} is loosely defined as a point that could be in $M$ under a slightly different mapping process. 

To solve the problem without ground-truth labels, a combinatorial optimization problem is defined over a sequential set of two point-cloud pairs $(\{M_0, S_0\}, \{M_1, S_1\})$. The first live scan is recorded at time $t_0$. Consider $t_1 = t_0 + \Delta t$ where $\Delta t$ is a hyperparameter of the system. All four point clouds are transformed into a common frame. The optimization has three terms: a chamfer loss \cite{chamfer} between the map and the \textit{Consistent} live scan, a class-balance loss that penalizes the number of points that are considered to be \textit{Changed}, and a temporal-consistency loss that penalizes labels that change between the sequential scans. The total loss function is defined as
\begin{equation}
\label{eq:totalLoss}
    \mathcal{L} = \mathcal{L}_{\text{cham}} + \lambda_1 \mathcal{L}_{\text{class}} + \lambda_2 \mathcal{L}_{\text{temporal}},
\end{equation}
where $\lambda_1$ and $\lambda_2$ are hyperparameters to be tuned. 
In theory, the solution to this problem does not require machine learning.
For a fixed set of maps and live-scans, a gradient descent method could be iterated to find a local minimum to the problem for each frame starting from random labels.
However, with thousands of points, it is intractable to solve as a classical optimization. To respect the run time requirements of a 10 Hz LiDAR scanner, a neural network is used to learn a fast heuristic solution to the optimization problem. 
To make the loss functions differentiable, the softmax probability of being classified as \textit{Changed} is used instead of the actual prediction. Next, each of the terms in \eqref{eq:totalLoss} is explained in detail.

\subsubsection{Chamfer Loss}
The chamfer loss uses the nearest-neighbour Euclidean distance between points in live scan ($s_i \in S$) and the map ($m_j \in M$). It is weighted by the likelihood that the given point belongs to the \textit{Consistent} class $(l_i = 0)$:
\begin{equation}
\label{eq:chamLoss}
    \mathcal{L}_{\text{cham}} = \frac{1}{n} \sum_{i=1}^n p(l_i = 0)\min_{m_j \in M} ||m_j - s_i||_2 .
\end{equation}
If the weighting of the chamfer loss is large, all points that do not exactly match the map will be classified as \textit{Changed}. The chamfer loss pushes points to be classified as \textit{Changed}.

\subsubsection{Class-Balance Loss}
The class-balance loss is the summation of the likelihood of a point being classified as \textit{Changed} $(l_i = 1)$:
\begin{equation}
\label{eq:classLoss}
    \mathcal{L}_{\text{class}} = \frac{1}{n} \sum_{i=1}^n p(l_i = 1) .
\end{equation}
This loss pushes all points to be classified as \textit{Consistent}. This loss opposes the chamfer loss.
 
\subsubsection{Temporal-Consistency Loss}
The temporal-consistency loss uses a bidirectional chamfer loss between points in the two consecutive live scans, $S_0$ and $S_1$, that are classified as \textit{Changed}.
$l_{ki}$ is the label of point $s_i$ in point cloud $S_k$:
\begin{equation}
\label{eq:tempLoss}
\begin{split}
    \mathcal{L}_{\text{temporal}} = \frac{1}{n_0} \sum_{i=1}^{n_0} p(l_{0i} = 1) \min_{s_{j} \in S_1} ||s_{i} - s_j||_2 \\
    + \frac{1}{n_1} \sum_{i=1}^{n_1} p(l_{1i} = 1) \min_{s_{j} \in S_0} ||s_i - s_j||_2 .
\end{split}
\end{equation}
This loss acts as a form of outlier rejection and encourages predictions to be consistent through time. This loss will push the network to classify all points as \textit{Consistent} because an empty obstacle class has a loss of zero.

\subsection{Deep Network Architecture}
The novelty of this paper lies in the training method rather than the network architecture used to process the 3D points. 
In \autoref{fig:blockArch}, the neural network could be any 3D architecture that can be modified to accept pairs of maps and live scans.
A convolutional neural network using range images as the input was chosen in this paper for its fast inference time \cite{milioto2019iros}.
Speed is critical to the autonomous operation of the UGV.
Preliminary experiments were also performed on a 3D architecture based on KPConv \cite{thomas_kpconv_2019} but the run time was too long.

\paragraph{Range Image Generation}
The inputs to the network are aligned range images from the map and the live scan. These images are rendered by transforming the live scan and local map 3D point clouds into a common spherical coordinate system centred on the LiDAR. 
Each range image pixel, $(u_i, v_i)$, contains the value $r_i = \sqrt{x_i^2+y_i^2+z_i^2}$, to the closest point in the frustum. 
Dimensions $H \times W = 64 \times 1024$ are used over a $25^\circ \times 360^\circ$ field of view (fov). 
The lower limit of the fov below the horizon ($\text{fov}_{\text{down}}$) is a sensor property.
An indexed mapping between the range image and point cloud is stored to reassign labels in the image to the corresponding 3D points.  
 Given a point $(x_i, y_i, z_i) \in \mathbb{R}^3$ the following image mapping $\mathbb{R}^3 \xrightarrow{} \mathbb{R}^2$ is used:
\begin{equation}
\label{eq:spherical}
\begin{pmatrix}
u_i\\
v_i\\
\end{pmatrix} = \\
\begin{pmatrix}
    \frac{1}{2}(1-\frac{\arctan(y_i, x_i)}{\pi})W \\
    \left(1-\frac{\arcsin(z_i, r_i) + \text{fov}_{\text{down}}}{\text{fov}} \right)H
\end{pmatrix}.
\end{equation}

\paragraph{Neural Network Architecture}
The range image network (RangeNetCD) is based on RangeNet++ \cite{milioto2019iros} but modified to accept both a map and live scan. The input has a dimension of $H \times W \times 2$ with the live scan and map on each channel.
The network encoder consists of four double convolutions followed by maxpooling. 
The decoder uses bilinear upsampling and skip connections from the encoder layers for per-pixel prediction. 
We observed that effective classification requires non-square convolution kernels on the first and last layers by experimenting with different shapes.
The kernel in the first and last layer has dimension $1\times 2$.

\section{Experiments}
Experimental evaluation occurred in forested and off-road areas around the University of Toronto Institute for Aerospace Studies (\autoref{fig:dataset}). A Clearpath Warthog UGV \cite{warthog} equipped with an Ouster-OS1 LiDAR \cite{ouster_128} was used for experiments. LiDAR ICP localization \cite{burnett_are_2022} aligned the live scan with the corresponding map prior to performing any neural-network inference. The horizon of the local planner is 10 m. Accordingly, detections are limited to this range around the robot.
Empirically, it was observed that the sensor-aligned frame with the origin on the ground below the LiDAR scanner performed the best.
This may occur because the nearby planar ground points are projected into the same pixel frustum. 

\begin{figure}
    \centering
    \includegraphics[scale=0.25]{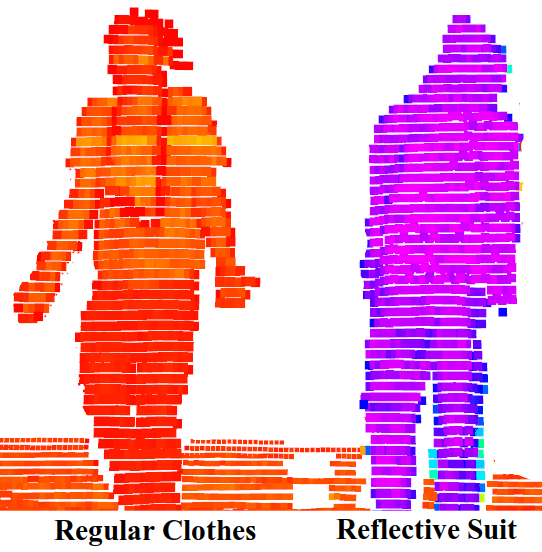}
    \caption{A LiDAR scan of two pedestrians coloured by intensity. Left: A pedestrian wearing regular clothes. Right: A pedestrian wearing the reflective suit.}
    \label{fig:coatPeople}
\end{figure}

\subsection{Retroreflective Semantic Evaluation}
While the method presented in this paper is fully unsupervised, for quantitative evaluation of the results, ground-truth labels are required. 
For change detection, an offline ray-tracing approach similar to Thomas et al. \cite{thomas_self-supervised_2020} could classify the points corresponding to changes of interest. However, these ray-tracing approaches are challenged in foliage \cite{wu_thesis} and are computationally expensive.
Instead, we propose using a rapid technique for implicitly labelling the data.

During dataset evaluation, all objects introduced into the scene are coated in retroreflective material. \autoref{fig:coatPeople} highlights the difference in intensity between a reflective suit, regular clothes, and the background.
The intensity channel of LiDAR is unique to each distinct sensor and not standardized between different models.
For this reason, we discard the intensity values and render the intensity channel out-of-band. 
However, the intensity can be used for automatic labelling: a threshold filter reliably classifies the points that belong to reflective objects.

\begin{table*}[h]
    \centering
    \caption{Statistics for the dataset used for training and quantitative evaluation}
    \begin{tabular}{|c|c|c|c|c|}
    \hline
    Sequence Name & Path Length (m) & Number of Traversals & Frames & Total Driving Time (h:mm) \\
    \hline
        Parking Loop & 925.7 & 5 & 11831 & 2:02\\
        \hline
        North Field Loop & 124.9 & 5 & 2250 & 0:18\\
        \hline
        Grass Yard & 338.7 & 3 & 2930 & 0:28\\
        \hline
        Forest Loop & 232.6 & 4 & 3557  & 0:33\\
        \hline
    \end{tabular}
    \label{tab:datasetStats}
\end{table*}

The effectiveness of this approach requires a controlled environment without ambient dynamic actors, so public scenarios are unsuitable.
However, for many cases when the rapid labelling of an object is required, this method is highly effective.
While we use pedestrians with reflective suits in our dataset, in general, retroreflective cloth or tape can be placed on any object where per-point ground truth is required.

\subsection{Results}
The network is trained and evaluated on a collection of four different mapping and repeating sequences. In total, the raw dataset consists of $50,000$ frames and the robot traverses $8.5$ km over $1.75$ km of unique paths.
A LiDAR frame is stored every 30 cm of driving rather than every 0.1 s, which reduces the total frames used to $20,568$.
This prevents biased training due to interruptions in the data collection when the robot stops moving.
The four sequences are traced out in \autoref{fig:dataset}.
The longest sequence is the Parking Loop, which is uncontrolled and contains moving vehicles, bicycles, and pedestrians.
The other three sequences were recorded with reflective obstacles for evaluation.
\autoref{tab:datasetStats} describes the details of each run of the dataset.

For comparison, a classical nearest-neighbour baseline \cite{girardeau2005change} is used as an alternative to unsupervised learning.
\autoref{tab:iou}, \autoref{tab:iou_med}, and \autoref{tab:iou_hard} provide the intersection-over-union (IoU) score of the \textit{Changed} class as well as the mean IoU (mIoU). Note that, because the labels are highly imbalanced, metrics evaluated on the \textit{Changed} class are the basis of comparison.
We adopt an additional planning-oriented metric to represent the ability of the robot to perform path planning.
The IoU is re-evaluated within the known planning corridor around the robot, which is 5 m wide for the Warthog running Teach and Repeat.
The results in Tables I-III are evaluated with map voxel size of 0.2 m and live-scan voxel size 0.05 m.
The loss function hyperparameters during training were $\lambda_1 = 15$ and $\lambda_2=1.0$.
\begin{table}[]
    \centering
    \caption{Performance on Easy (Grass Yard) data of unsupervised detector and classical baseline}
    \begin{tabular}{|c|c|c|c|c|}
    \hline
    Method & $\text{IoU}_{\text{ch}}$ & Corridor $\text{IoU}_{\text{ch}}$ & mIoU \\
    \hline
        Nearest Neighbour & 0.507 & 0.618 & 0.745\\
        \hline
        RangeNetCD (ours) & \textbf{0.651} & \textbf{0.820} & \textbf{0.822}\\
        \hline
    \end{tabular}
    \label{tab:iou}
\end{table}

\begin{table}[]
    \centering
    \caption{Performance on Medium (North Field) data of unsupervised detector and classical baseline}
    \begin{tabular}{|c|c|c|c|c|}
    \hline
    Method & $\text{IoU}_{\text{ch}}$ & Corridor $\text{IoU}_{\text{ch}}$ & mIoU \\
    \hline
        Nearest Neighbour & 0.438 & 0.586 & 0.712\\
        \hline
        RangeNetCD (ours) & \textbf{0.577} & \textbf{0.609} & \textbf{0.782}\\
        \hline
    \end{tabular}
    \label{tab:iou_med}
\end{table}

\begin{table}[]
    \centering
    \caption{Performance on Hard (Forest Loop) data of unsupervised detector and classical baseline}
    \begin{tabular}{|c|c|c|c|c|}
    \hline
    Method & $\text{IoU}_{\text{ch}}$ & Corridor $\text{IoU}_{\text{ch}}$ & mIoU \\
    \hline
        Nearest Neighbour & 0.290 & 0.547 & 0.635\\
        \hline
        RangeNetCD (ours) & \textbf{0.607} & \textbf{0.656} & \textbf{0.673}\\
        \hline
    \end{tabular}
    \label{tab:iou_hard}
\end{table}

On the Grass Yard (\autoref{tab:iou}), the corridor $\text{IoU}_{\text{ch}}$ is 16.9\% higher than the $\text{IoU}_\text{ch}$ evaluated on the entire scan for RangeNetCD.
This highlights the practical effectiveness of the system because false positives on static structures outside the corridor of the planner do not affect the trajectory. 

\begin{figure}
    \centering
    \includegraphics[scale=0.6]{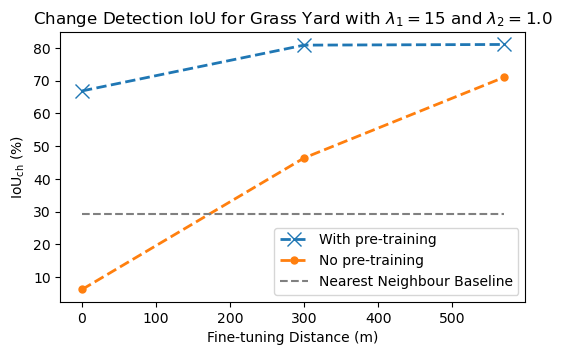}
    \caption{Changed IoU vs Distance Traveled Fine-tuning for a map-voxel of 0.3 m and live voxel of 0.05 m. Pre-training improves the performance.}
    \label{fig:train_plot}
\end{figure}

We find that pre-training a general network followed by environment-specific fine-tuning is the most effective training approach. 
To pre-train, the unlabelled and uncontrolled data from the Parking Loop is used.
Finetuning occurs on a specific sequence with a lower learning rate in the optimizer. In \autoref{fig:train_plot}, pre-training improves the initial performance to 68\%.
While the general network is capable in all of the environments tested, fine-tuning with more specific examples boosts detection accuracy further. 
An avenue of future work is to fine-tune live on the robot once it has been deployed in operation.
The detection quality is suitable for autonomous navigation. 

\subsection{Ablation Studies}

\paragraph{Scan and Map Density}
The OS-1 LiDAR used for these experiments produces 1.3 million points per second \cite{ouster_128}. Using every point is computationally impractical for ICP odometry and localization \cite{wu_thesis}.
Additionally, storing every point in the map is prohibitively expensive. For this reason, the map is voxel downsampled.
The voxel sizes of the map and live scan impact the accuracy of change detection.
\autoref{fig:mars_study} shows the corridor IoU of the \textit{Changed} class for different combinations of scan and map densities.
Increasing the live-scan density improves performance but changing the map density is less important. Most obstacles in the dataset are smaller than 1 m in their longest dimension so voxel downsampling removes defining points. 
In contrast, much of the map contains larger objects that can be downsampled without affecting performance. 
The range-image network performs much better than the baseline as the map density decreases. This is advantageous because smaller point clouds use less storage and are processed faster.
For reference, storing the raw scan requires about 50 Gb/km and downsampling to 0.3 m requires about 3.75 Gb/km. Intermediate values scale nonlinearly based on the LiDAR scan pattern and density of nearby static features.
A map voxel size of 0.3 m and a live-scan voxel size of 0.05 m is used on the Warthog.

\begin{table}[]
\caption{Corridor $\text{IoU}_{\text{ch}}$ for different combinations of the map and live scan voxel downsampling on the Grass Yard}
\label{fig:mars_study}

\centering
\begin{tabular}{cc|ccc|}
\cline{3-5}
                                                           &      & \multicolumn{3}{c|}{Map Voxel Size}                   \\ \cline{3-5} 
                                                           &      & \multicolumn{1}{c|}{0.1 m} & \multicolumn{1}{c|}{0.2 m} & \multicolumn{1}{c|}{0.3 m} \\ \hline
\multicolumn{1}{|c|}{\multirow{3}{*}{Live Voxel Size}} & 0.05 m & \multicolumn{1}{c|}{0.760} & \multicolumn{1}{c|}{\textbf{0.820}} & \multicolumn{1}{c|}{0.811}  \\ \cline{2-5} 
\multicolumn{1}{|c|}{}                                     & 0.15 m & \multicolumn{1}{c|}{0.091} & \multicolumn{1}{c|}{0.111} &  \multicolumn{1}{c|}{0.110}   \\ \cline{2-5} 
\multicolumn{1}{|c|}{}                                     & 0.3 m  & \multicolumn{1}{c|}{0.050}& \multicolumn{1}{c|}{0.071}    &  \multicolumn{1}{c|}{0.056}   \\ \hline
\end{tabular}
\end{table}

\paragraph{Loss Functions}
\autoref{tab:loss_ablate} shows the impact of each loss function on the performance for the 0.2 m and 0.05 m map and live-scan voxelizations of the Grass Yard.
As expected, if the class loss is removed, all points are classified as \textit{Changed}. This occurs because the trivial optimal solution classifies all points as \textit{Changed}.
Conversely, if only the class loss is used, all points are classified as \textit{Consistent}. 
The correct balance of the chamfer and class losses accounts for the basic capability of the system.
Adding the temporal loss improves the performance by penalizing transient points, reducing the number of false-positive detections. 
\begin{table}[]
    \caption{Ablation Study on Loss Functions evaluated on the Grass Yard with Map Voxel 0.2 m and Live Voxel 0.05 m}
    \centering    
    \begin{tabular}{|c|c|c|c|}
    \hline
        Chamfer & Class & Temporal & $\text{IoU}_{\text{ch}}$ \\
        \hline
        $\checkmark$ & $\times$ & $\times$ & 0.02 \\
        \hline
        $\times$ & $\checkmark$ & $\times$ & 0.0 \\
        \hline
        $\checkmark$ & $\checkmark$ & $\times$ & 0.388 \\
        \hline
        $\checkmark$ & $\checkmark$ & $\checkmark$ & \textbf{ 0.651} \\
        \hline
    \end{tabular}
    \label{tab:loss_ablate}
\end{table}

\subsection{Qualitative Evaluation}

\begin{figure}
    \centering
    \includegraphics[width=0.5\textwidth]{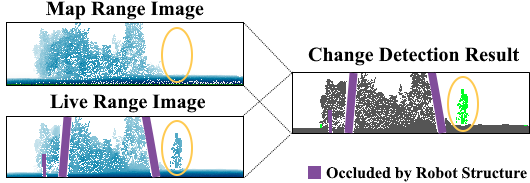}
    \caption{The input range images and change detection of a cyclist passing by the robot. The two range images are used by the network to determine changes. Part of the live scan is blocked by the robot's structure, this area is purple. In the result, green regions are classified as changed and grey regions are consistent.}
    \label{fig:cyclist}
\end{figure}

Several interesting phenomena are observed in the output of the network.
First, this change-detection approach is capable of generalizing to new types of objects. In \autoref{fig:cyclist}, a cyclist is accurately detected biking past the robot. No cyclists were included in the training data.
While RangeNetCD is designed to allow for the detection of any changes in the scene, it is possible to overfit the network to shapes seen during training. For example, when the network was trained on data that changed only by introducing pedestrians, it failed to detect new vehicles at test time.
This was solved by using more, diverse training data.
With extensive datasets, it is interesting to consider if these objects could be differentiated in the feature space.
We observe that if training sequences contain changes in every frame, the network forces detections in every frame. The most false positives appear when the correct solution is no change at all.

Change-detection performance is limited by the map quality. The map is assumed to capture only the permanent environment but errors appear when it does not. 
Using the uncontrolled parking lot data, a map was generated containing cars that had moved before later runs were recorded. This led to two types of misdetections. Points on static objects that were previously occluded are false positives. For example, a fire hydrant behind a parked car was later detected as a change even though it was always present.
Points in the same location as an object that moved are false negatives. For example, a pedestrian walking through an empty parking space where a car was located during mapping is not detected even though it is a change. 
These issues rarely impact the planner, because the taught path will not exist inside obstacles in the map.
Significant issues with the mapping process can destabilize network training because patterns are less consistent. 
These errors were eliminated by generating the map when the parking lots were empty. 
Applying a map-cleaning approach offline would be beneficial as well.

\subsection{Closed-Loop Performance}
The range-image network was tested on a laptop NVIDIA RTX A4500 GPU located on the robot.
Inference takes $14.9 \pm  2.3$ ms. 
Most of this time is spent transferring the tensors between CPU and GPU memory. More optimal pipelines may exist that reduce the amount of transfer.
RangeNetCD was exported to TorchScript and executed in C++ as part of the existing LiDAR Teach and Repeat \cite{wu_thesis} pipeline. 

\begin{figure}
    \centering
    \includegraphics[width=0.49\textwidth]{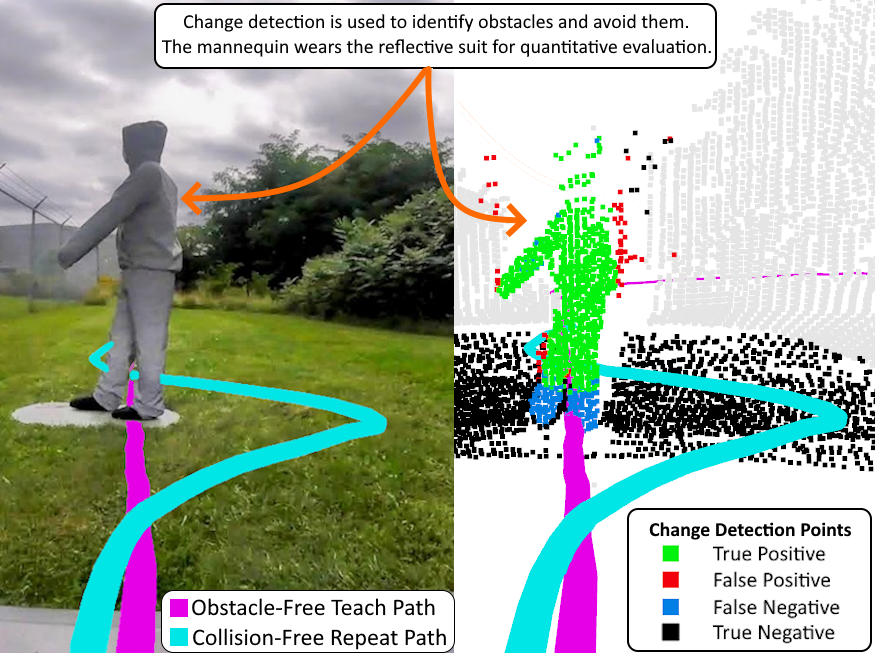}
    \caption{Left: an image view of an obstacle and the path the robot drives to avoid it. Right: The point cloud view of the same scene.}
    \label{fig:sideBySide}
\end{figure}

Points detected as \textit{Changed} are passed to a cost map inflation module that projects them into 2D and accounts for the robot's radius. 
The planner maintains a queue of cost maps to overcome the LiDAR's close-range blind spot. 
These cost maps allow the robot to avoid local obstacles. 
The supplementary video \footnote{\href{https://youtu.be/prqVQJHXYWE}{https://youtu.be/prqVQJHXYWE}} contains sequences of the robot navigating around a series of introduced obstacles. 
In \autoref{fig:sideBySide}, a mannequin was placed on the path as a change. The robot successfully detects and avoids it and then continues to follow the originally planned path. 
As a side effect of maintaining a queue of obstacles, dynamic changes cause smearing in the cost map leading to suboptimal routes. As a future extension to this work, a self-supervised prediction layer could be used to extrapolate the motion of changes over a short time horizon. 

\section{Conclusion and Future Work}
In this paper, we present a novel method for the unsupervised training of a deep network that performs change detection on a pair of point clouds. 
We show that by exploiting inductive biases related to the amount of change, distance of changes from the map, and temporal consistency of physical scenes, it is possible to train a network. 
We demonstrate the applicability of the approach on a range-image convolutional neural network that is trained on data collected on an unmanned ground vehicle. 
Once trained, the network runs quickly and is added to the robot to improve its ability to navigate for long periods.
Future work using a pre-trained contrastive encoder for feature extraction will add semantic differences to the training loss.

\section*{Acknowledgment}
We thank Nikhil Thiyagarajan for his assistance in collecting the dataset.
This work was supported by the Natural Sciences and Engineering Research Council of Canada (NSERC) and the Vector Scholarship for AI. 

\bibliographystyle{ieeetr}
\bibliography{refs}

\end{document}